# Quranic Conversations: Developing a Semantic Search tool for the Quran using Arabic NLP Techniques

Yasser Shohoud

Senior Software Engineering Manager
Meta Facebook
Menlo Park, California, United States
ORCID: 0009-0008-9141-4343

Maged Shoman

Postdoctoral Research Associate
University of British Columbia
Vancouver, BC V6T 1Z4
maged.shoman@ubc.ca
ORCID: 0000-0002-4265-071X

Sarah Abdelazim

Data Scientist
University of British Columbia
Vancouver, BC V6T 1Z4
missarah@student.ubc.ca
ORCID: 0000-0002-6379-2733

# Abstract

The Holy Book of Qur'an is believed to be the literal word of God (Allah) as revealed to the Prophet Muhammad (PBUH) over a period of approximately 23 years. It is the book where God provides guidance on how to live a righteous and just life, emphasizing principles like honesty, compassion, charity and justice, as well as providing rules for personal conduct, family matters, business ethics and much more. However, due to constraints related to the language and the Qur'an organization, it is challenging for Muslims to get all relevant ayahs (verses) pertaining to a matter or inquiry of interest. Hence, we developed a Qur'an semantic search tool which finds the verses pertaining to the user inquiry or prompt. To achieve this, we trained several models on a large dataset of over 30 tafsirs, where typically each tafsir corresponds to one verse in the Qur'an and, using cosine similarity, obtained the tafsir tensor which is most similar to the prompt tensor of interest, which was then used to index for the corresponding ayah in the Qur'an. Using the SNxLM model, we were able to achieve a cosine similarity score as high as 0.97 which corresponds to the "abdu" tafsir for a verse relating to financial matters.

# Introduction

The world's Muslim community is estimated at two billion Muslims today and expected to reach 2.2 billion by 2030 [1]. Being a Muslim entails having faith in the Qur'an as the divine word of God, which was revealed to Prophet Muhammad (PBUH) more than fourteen centuries ago. Consequently, it is the foundation of faith, the primary source of Islamic law, and the source of truth [2]. Muslims believe that Prophet Muhammad received the Qur'an in revelations over a 22-year period beginning in 610 CE and ending with his death in 632 CE. The Prophet would recite the revelation to his companions who would memorize it and write it on various available writing surfaces such as animal skins or bones. During this period, the Prophet ensured that several of his followers had memorized the Qur'an and could recite it correctly. One year after the Prophet's death, Abu Bakr appointed Zayd Ibn Thabit to produce a complete written copy of the Qur'an. Zayd Ibn Thabit, who had memorized the Qur'an and served as a personal secretary to the Prophet, produced a copy of the Qur'an that was passed down to Abu Bakr's daughter and Prophet's wife, Hafsah. In 644, the Muslim empire was rapidly growing, creating a need for standardized copies of the Qur'an that can be sent to all corners of the empire to help educate new Muslims in their religion. In 644 CE, the third Caliph 'Uthman ibn 'Affan directed Zayd Ibn Thabit to utilize the copy of the Qur'an, which he had authored and entrusted to Hafsah, to make a final recension of the Qur'an. This effort aimed to standardize and resolve minor variations in the Arabic dialects spoken during that period [3]. The Qur'an has also been transmitted generationally via memorization. There are tens of thousands of Muslims living today who have memorized all 6,236 ayahs (verses) of the Qur'an, learning it from teachers who can trace back the chain of transmission to the Prophet. The veracity of the Qur'an draws from the early standardized writing and multiple traceable chains of transmission making it the reliable, trusted source of truth for Muslims.

Muslims frequently seek guidance from the Qur'an regarding various topics or questions as they navigate their lives. This serves as an initial step in comprehending the Islamic perspective on addressing these matters. Despite the abundance of books, websites, and apps offering the Arabic Qur'an and its translations, achieving reliable access to these resources today is surprisingly challenging due to various reasons. First, keyword search is insufficient because there are many relevant ayahs to a topic that may not explicitly mention that topic by name. For example, when searching for the word Adam the results do not include ayah 2:30:

> [Prophet], when your Lord told the angels, ‘I am putting a successor on earth’, they said, ‘How can You put someone there who will cause damage and bloodshed, when we celebrate Your praise and proclaim Your holiness?’ but He said, ‘I know things you do not.’[4]

Second, there is no authoritative, widely agreed upon, topic index for the Qur’an that Muslims can use to find all ayahs related to a topic of interest. Third, ayahs related to the topic of interest may use different words to convey specific segments or subtopics. For example, the words *nabi* and *rasūl* are used to distinguish between prophet and messenger, and *zakāh* and *sadaqah* are used to distinguish between obligatory alms and voluntary charity. Finally, the Muslim tradition allows differences of opinion in the meaning of ayahs and therefore their relevance to a topic of interest. For example, in ayah 111:4 the phrase *hammālat al-hattab* is usually translated as the firewood-carrier [5] but, according to Lane’s Lexicon and Gharib Al Qur’an, it also means the one who incites, urges or instigates people against someone [6]. Given the existing constraints, our aim is to develop a semantic search tool for the Qur'an which enables Muslims to efficiently locate all pertinent verses related to a specific topic or question, allowing them to accurately address the query, "What does the Qur'an say about ____ ?"

# Methodology

## Quran and Topics Index

We used the Hafs reading of the Qur’an as obtained in plain text from King Fahd Glorious Qur’an Complex [7]. There is no canonical Quran topic index so we combined the indexes from SearchTruth [8] and the Quranic Corpus [9] to form a unified index of topics.

## Quran Tafsirs

We used thirty Arabic tafsirs (exegesis) and two English commentaries obtained from Qortoba Foundation [10]. We used tafsirs from the Sunni, Shia, and Sufi schools and a variety of periods ranging from the tenth to the twenty-first century with different exegetical approaches. Some of the tafsirs, like Al-Jalalayn [11], provide succinct commentary on each verse separately making it easy to map the text of the commentary to a specific verse. The commentary in other tafsirs, like Al Tafsir Al Kabir [12], spans multiple sequential verses that together cover a topic or theme so we mapped such commentary to each of the verses in the sequence.

## Tafsir Methodologies

Tafsir methodologies are often divided into two broad categories: Tafsir bi'l-Ma'thur and Tafsir bi'r-Ra'y.

Tafsir bi'l-Ma'thur (Transmitted Interpretation): Tafsir bi'l-Ma'thur methodology relies on established sources of Muslim tradition including the Qur’an itself, sayings of the Prophet (PBUH), and sayings of his companions and followers. Among those sources, the Qur’an is the most authoritative and it is common for exegetes to use Qur’an verses to explain or clarify another verse and sometimes to explain the meaning of a specific word. Tafsirs following this methodology often focus on the Arabic language, grammatical analysis, and sometimes poetry, to explain word and phrase meanings.

*Tafsir bi'r-Ra'y (Intellectual Interpretation): The scarcity of traditional sources available has led some exegetes to apply independent rational reasoning and personal intellect to construct an interpretation based on individual opinions. This method, known as tafsir bi'r-ra'y, is distinguished by the incorporation of the commentator's viewpoints, thereby fostering a diverse and balanced comprehension of Quranic verses. This approach finds support within the Quran itself, as indicated in Surah Sad (Qur’an chapter 38) verse 29:*

> This is a Scripture that We have revealed to you, abounding in blessings, so that they may ponder its verses and that people of understanding may reflect.

However, it should be noted that this method does not advocate a subjective interpretation solely based on personal opinions. Rather, opinions should be rooted in the primary sources and the Arabic language.

## Examples from Tafsir

In this section we provide a brief description of each tafsir followed by an example of the tafsir and an English translation, translated using ChatGPT [14].

### *Al-Kashshaaf 'an Haqa'iq at-Tanzil by Al-Zamakhshari*

Written by the 12th century Mu'tazilite scholar al-Zamakhshari, Al-Kashshaaf 'an Haqa'iq at-Tanzil, popularly known as Al-Kashshaaf i.e. 'the Revealer' has commanded a lot of respect and recognition due to its comprehensive analysis of language and its exposition of meanings conveyed through the Quran's literary devices and figurative language. It has also faced criticism for incorporating philosophical viewpoints influenced by Mu'tazilah thought. Throughout the exegesis, the author uses a conversational question and answer approach where he begins the question by "if you say" and follows it with an answer beginning by "I would say". The author divides the tafsir in sequences of related verses rather than one verse at a time. In some cases, the sequence can be the entire surah. As an example, consider the tafsir of the surah 102:

أَلْهَىٰكُمُ ٱلتَّكَاثُرُ ١ حَتَّىٰ زُرْتُمُ ٱلْمَقَابِرَ ٢ كَلَّا سَوْفَ تَعْلَمُونَ ٣ ثُمَّ كَلَّا سَوْفَ تَعْلَمُونَ ٤ كَلَّا لَوْ تَعْلَمُونَ عِلْمَ ٱلْيَقِينِ
٥ لَتَرَوُنَّ ٱلْجَحِيمَ ٦ ثُمَّ لَتَرَوُنَّهَا عَيْنَ ٱلْيَقِينِ ٧ ثُمَّ لَتُسْـَٔلُنَّ يَوْمَئِذٍ عَنِ ٱلنَّعِيمِ ٨

> 1 Competing for more distracts you 2 until you go into your graves. 3 No indeed! You will come to know. 4 No indeed! In the end you will come to know. 5 No indeed! If only you knew for certain! 6 You will most definitely see Hellfire. 7 Then you will see it with the eye of certainty. 8 On that Day, you will be asked about your pleasures [13].

The Al-Kashshaaf tafsir is for the entire surah in one continuous narrative whereas other tafsirs address each verse separately. Here is the Al-Kashshaaf tafsir in Arabic followed by the English translation.

ألهاه عن كذا وأقهاه: إذا شغله. و ﴿التكاثر﴾ التباري في الكثرة والتباهي بها، وأن يقول هؤلاء: نحن أكثر، وهؤلاء: نحن أكثر. روي أن بني عبد مناف وبني سهم تفاخروا أيهم أكثر عدداً، فكثرهم بنو عبد مناف فقالت بنو سهم: إن البغي أهلكنا في الجاهلية فعادّونا بالأحياء والأموات، فكثرتهم بنو سهم. والمعنى: أنكم تكاثرتم بالأحياء حتى إذا استوعبتم عددهم صرتم إلى المقابر فتكاثرتم بالأموات: عبر عن بلوغهم ذكر الموتى بزيارة المقابر تهكماً بهم. وقيل: كانوا يزورون المقابر فيقولون: هذا قبر فلان وهذا قبر فلان عند تفاخرهم. والمعنى: ألهاكم ذلك- وهو مما لا يعينكم ولا يجدي عليكم في دنياكم وآخرتكم- عما يعينكم من أمر الدين الذي هو أهم وأعني من كل مهم. أو أراد ألهاكم التكاثر بالأموال والأولاد إلى أن متم وقبرتم. منفقين أعماركم في طلب الدنيا والاستباق إليها والتهالك عليها، إلى أن أتاكم الموت لا همّ لكم غيرها، عما هو أولى بكم من السعي لعاقبتكم والعمل لآخرتكم. وزيارة القبور: عبارة عن الموت. قال:

لَنْ يُخْلِصَ الْعَامَ خَلِيلٌ عِشْراً ** ذَاقَ الضِّمَادَ أَوْ يَزُورَ الْقَبْرا
وقال:

زَارَ الْقُبُورَ أَبُو مَالِكٍ ** فَأَصْبَحَ أَلأَمَّ زُوَّارِهَا
وقرأ ابن عباس: ﴿أألهاكم﴾ ؟ على الاستفهام الذي معناه التقرير ﴿كَلاَّ﴾ ردع وتنبيه على أنه لا ينبغي للناظر لنفسه أن تكون الدنيا جميع همه ولا يهتم بدينه ﴿سَوْفَ تَعْلَمُونَ﴾ إنذار ليخافوا فيتنبهوا من غفلتهم. والتكرير: تأكيد للردع والأنذار عليهم. و(ثم) دلالة على أن الإنذار الثاني أبلغ من الأوّل وأشد، كما تقول للمنصوح: أقول لك ثم أقول لك: لا تفعل، والمعنى: سوف تعلمون الخطأ فيما أنتم عليه إذا عاينتم ما قدّامكم من هول لقاء الله، وإنّ هذا التنبيه نصيحة لكم ورحمة عليكم. ثم كرّر التنبيه أيضاً وقال: ﴿لَّوْ تَعْلَمُونَ﴾ محذوف الجواب، يعني: لو تعلمون ما بين أيديكم علم الأمر اليقين، أي: كعلمكم ما تستيقنونه من الأمور التي وكلتم بعملها هممكم: لفعلتم ما لا يوصف ولا يكتنه؛ ولكنكم ضلال جهلة؛ ثم قال: ﴿لَتَرَوُنَّ الجحيم (6)﴾ فبين لهم ما أنذرهم منه وأوعدهم به؛ وقد مرّ ما في إيضاح الشيء بعد إبهامه من تفخيمه وتعظيمه، وهو جواب قسم محذوف، والقسم لتوكيد الوعيد، وأن ما أوعدوا به ما لا مدخل فيه للريب؛ وكرره معطوفاً بثم تغليظاً في التهديد وزيادة في التهويل. وقرئ: ﴿لترؤن﴾ بالهمز وهي مستكرهة.

فإن قلت: لم استكرهت والواو المضمومة قَلْبَهَا همزة قياس مطرد؟ قلت: ذاك في الواو التي ضمتها لازمة، وهذه عارضة لالتقاء الساكنين. وقرئ: ﴿لترون﴾ ولترونها: على البناء للمفعول ﴿عَيْنَ اليقين﴾ أي: الرؤية التي هي نفس اليقين وخالصته.

ويجوز أن يراد بالرؤية: العلم والإبصار ﴿عَنِ النعيم﴾ عن اللهو والتنعم الذي شغلكم الالتذاذ به عن الدين وتكاليفه.

فإن قلت: ما النعيم الذي يسئل عنه الإنسان ويعاتب عليه؟ فما من أحد إلاّ وله نعيم؟ قلت: هو نعيم من عكف همته على استيفاء اللذات، ولم يعش إلاّ ليأكل الطيب ويلبس اللين، ويقطع أوقاته باللهو والطرب، لا يعبأ بالعلم والعمل، ولا يحمّل نفسه مشاقهما؛ فأما من تمتع بنعمة الله وأرزاقه التي لم يخلقها إلاّ لعباده، وتقوّى بها على دراسة العلم والقيام بالعمل، وكان ناهضاً بالشكر: فهو من ذاك بمعزل؛ وإليه أشار رسول الله صلى الله عليه وسلم فيما يروي: أنه أكل هو وأصحابه تمر وشربوا عليه ماء فقال: «الحمد لله الذي أطعمنا وسقانا وجعلنا مسلمين»

عن رسول الله صلى الله عليه وسلم: «من قرأ ألهاكم التكاثر لم يحاسبه الله بالنعيم الذي أنعم به عليه في دار الدنيا، وأعطي من الأجر كأنما قرأ ألف آية».

It preoccupied them with such matters and diverted them. And "Al-Takaathur" signifies competing in abundance and boasting about it, as if some say: "We are more numerous," and others say: "We are more numerous." It is narrated that the descendants of 'Abd Manaf and the descendants of Sahm boasted about who had a greater number. The descendants of 'Abd Manaf boasted about their numbers, so the descendants of Sahm said, "In the era of ignorance, injustice destroyed us, and they retaliated against us with the living and the dead," so the descendants of Sahm boasted about their numbers. The meaning is that you have boasted about the number of your living until you have encompassed their number, but then you turned towards the graves and boasted about the number of your dead. This is an expression indicating their mentioning of the dead by visiting the graves, as a form of boasting. It is also said that they used to visit the graves and say, "This is the grave of so-and-so, and this is the grave of so-and-so," as part of their boasting. The meaning is that this matter, which does not benefit you and does not yield any results in your worldly life or the hereafter, has preoccupied you from the important matters of religion, which are more significant and more beneficial than everything else. Or it may mean that the pursuit of accumulating wealth and children has preoccupied you until your death and burial. You have spent your lives seeking worldly possessions, racing towards them, and exhausting yourselves for them until death comes to you, with no concern other than that. It has distracted you from that which is more deserving of your efforts, namely striving for your ultimate fate and working for your afterlife. And visiting graves is a metaphor for death. It is said:

The passing year does not spare even the closest friends,
They taste the shroud or visit the grave.
And it is also said:
Abu Malik visited the graves,
And in the morning, he became the one visited by pain.

Ibn Abbas recited: "Has the worldly life deluded you and preoccupied you?" as an interrogative statement that implies affirmation. "Nay, you will come to know!" It serves as a warning and a reminder that it is not appropriate for a person to have the worldly life as their sole concern and neglect their religion. "Soon you will come to know" serves as a warning for them to fear and be cautious of their negligence. The repetition is a reinforcement of the warning and admonition towards them. And "thumma" (then) indicates that the second warning is more severe and intense, as if you say to someone you advise, "I tell you, and then I tell you: Do not do it." The meaning is that you will come to realize the error of your ways when you witness the magnitude of facing Allah. This warning is an advice and mercy for you. Then the warning is repeated again, saying: "If only you knew." The answer is omitted, meaning: If only you knew the certain knowledge that lies before you, meaning what you are absolutely certain about, you would not engage in futile actions and deeds. But you are misguided and ignorant. Then it is stated: "Indeed, you will surely see the Hellfire." It clarifies what they have been warned about and what has been promised to them. It is similar to what was mentioned earlier, emphasizing its significance and magnifying its importance. It is a response to an omitted oath, and the oath is for the sake of emphasizing the threat and ensuring that what they have been promised is beyond doubt. It is repeated again, followed by an intensification of the threat and an increase in exaggeration. And it is recited as "latarawunnaha" with a hamzah, which is an alternative recitation.

If you say, "I did not notice it, and the suppressed 'waw' is changed to a hamzah due to continuous pattern (qiyas)," I say: That is in regard to the 'waw' that is connected to the context, while this is an indication of the meeting of two consonants. And it is recited as "laturoon" and "laturoonaha," using the structure for the object. "The eye of certainty" means the vision that is equivalent to certainty itself, pure and unadulterated.

It is also possible that by "seeing," it refers to knowledge and insight. "From the pleasures" means turning away from idle enjoyment and indulgence that have distracted you from religion and its obligations.

If you ask, "What pleasure is the human being inquiring about and reproaching himself for?" Isn't everyone blessed with pleasures? I say: It refers to the pleasure of one who dedicates their efforts solely to fulfill desires, living only to consume delicacies and wear luxuries, wasting their time in amusement and entertainment, disregarding knowledge and deeds, and avoiding any hardship associated with them. As for the one who enjoys the blessings and provisions from Allah, which He created solely for His servants, and utilizes them to study knowledge and engage in righteous deeds, displaying gratitude, they fall into the aforementioned category. The Prophet Muhammad, peace be upon him, indicated this in what is narrated, that he and his companions ate dates and drank water, and he said, "Praise be to Allah, who has fed us, given us drink, and made us Muslims."

From the Messenger of Allah, peace be upon him: "Whoever recites 'Has the worldly life deluded you?' will not be held accountable by Allah for the pleasures that were bestowed upon him in the worldly abode, and he will receive reward as if he had recited a thousand verses."

***Al-Mukhtasar by Tafsir Center for Qur'anic Studies***

Published in 2013, this modern tafsir aims to explain the meaning of the Quran verse-by-verse in a short and easy to understand manner. It is the result of a collaboration of twenty scholars and editors and relies heavily on transmitted interpretations (tafsir bi'l-ma'thur) while selecting the most common interpretation and summarizing it to one or a few sentences. By contrast to Al-Kashshaf, this tafsir separates each verse individually facilitating the creation of verse embeddings specific to the context of the verse.

For example, here is the tafsir for the first verse of surah 102:

أَلْهَاكُمُ التَّكَاثُرُ

شغلكم - أيها الناس - التفاخر بالأموال والأولاد عن طاعة الله.

You are preoccupied - O people - with boasting about wealth and children rather than obedience to God.

### *Muharar al-Wajiz by Ibn Atiyah al-Andalusi*

Ibn Attiyah combined the tafsir bi'l-ma'thur and tafsir bi'r-ra'y aspects of Quran exegesis as he often mentions what has been narrated from the Prophet Muhammad (PBUH), his companions, and the followers. He also extensively discusses the various possibilities of interpreting the verses, citing the opinions of other commentators and exegetes. The author relies on the Arabic language to clarify some meanings, often citing Arabic poetry, literary examples, and grammatical analysis. This tafsir, like Al-Kashshaf, addresses sequences of related verses that can sometimes include the entire surah. Example tafsir for surah 102:

بِسْمِ اللهِ الرَحْمَنِ الرَحِيمِ
تَفْسِيرُ سُورَةِ التَكاثُرِ
وهِيَ مَكِّيَّةٌ لا أعْلَمُ فِيها خِلافًا.
قوله عزّ وجلّ:

﴿
أَلْهَىٰكُمُ ٱلتَّكَاثُرُ ١ حَتَّىٰ زُرْتُمُ ٱلْمَقَابِرَ ٢ كَلَّا سَوْفَ تَعْلَمُونَ ٣ ثُمَّ كَلَّا سَوْفَ تَعْلَمُونَ ٤ كَلَّا لَوْ تَعْلَمُونَ عِلْمَ ٱلْيَقِينِ
٥ لَتَرَوُنَّ ٱلْجَحِيمَ ٦ ثُمَّ لَتَرَوُنَّهَا عَيْنَ ٱلْيَقِينِ ٧ ثُمَّ لَتُسْـَٔلُنَّ يَوْمَئِذٍ عَنِ ٱلنَّعِيمِ ٨

"ألْهاكُمْ" مَعْناهُ: شَغَلَكم بِلَذّاتِهِ، ومِنهُ "لَهْوُ الحَدِيثِ والأصْواتِ" واللَهْوُ بِالنِساءِ، وهَذا خَبَرٌ فِيهِ تَقْرِيعٌ وتَوْبِيخٌ وتَحَسُّرٌ. وقَرَأ ابْنُ عَبّاسٍ وأبُو عِمْرانَ الجَوْنِيُّ، وأبُو صالِحٍ: "آلْهاكُمْ" عَلى الِاسْتِفْهامِ.
و"التَكاثُرُ" هو المُفاخَرَةُ بِالأمْوالِ والأولادِ والعَدَدِ جُمْلَةً، وهَذا هِجِّيرى أهْلِ الدُنْيا وأبْنائِها العَرَبِ وغَيْرِهِمْ، لا يَتَخَلَّصُ مِنهُ إلّا العُلَماءُ المُتَّقُونَ، وقَدْ قالَ الأعْشى:

ولِسْتُ بِالأكْثَرِ مِنهم حَصًى وإنَّما العِزَّةُ لِلْكاثِرِ

وقالَ النَبِيُّ : « "يَقُولُ ابْنُ آدَمَ: مالِي مالِي، وهَلْ لَكَ مِن مالِكَ إلّا ما أكَلْتَ فَأفْنَيْتَ، أو لَبِسْتَ فَأبْلَيْتَ، أو تَصَدَّقْتَ فَأمْضَيْتَ"؟».
واخْتَلَفَ المُتَأوِّلُونَ في مَعْنى قَوْلِهِ تَعالى: ﴿حَتّى زُرْتُمُ المَقابِرَ﴾، فَقالَ بَعْضُهُمْ: حَتّى ذَكَرْتُمُ المَوْتى في تَفاخُرِكم بِالآباءِ والسَلَفِ، وتَكَثَّرْتُمْ بِالعِظامِ الرَمِيمِ، وقالَ آخَرُونَ: المَعْنى: حَتّى مُتُّمْ وزُرْتُمْ بِأجْسادِكم مَقابِرَكُمْ، أيْ قَطَعْتُمْ بِالتَكاثُرِ أعْمارَهُمْ، وعَلى هَذا التَأْوِيلِ رُوِيَ أنَّ أعْرابِيًّا سَمِعَ هَذِهِ الآيَةَ فَقالَ: بُعِثَ القَوْمُ لِلْقِيامَةِ ورَبِّ الكَعْبَةِ، فَإنَّ الزائِرَ مُنْصَرِفٌ لا يُقِيمُ، وحَكى النَقّاشُ هَذِهِ النَزْعَةَ عن عُمَرَ بْنِ عَبْدِ العَزِيزِ، وقالَ آخَرُونَ: هَذا تَأْنِيبٌ عَلى الإكْثارِ مِن زِيارَةِ القُبُورِ، أيْ: جَعَلْتُمْ أشْغالَكُمُ القاطِعَةَ لَكم عَنِ العِلْمِ والتَعَلُّمِ زِيارَةُ القُبُورِ تَكَثُّرًا بِمَن سَلَفَ وإشادَةً بِذِكْرِهِ، وقالَ: ثُمَّ قالَ النَبِيُّ عَلَيْهِ الصَلاةُ والسَلامُ: « "كُنْتُ نَهَيْتُكم عن زِيارَةِ القُبُورِ فَزُورُوها ولا تَقُولُوا هَجْرًا".» فَكانَ نَهْيُهُ عَلَيْهِ الصَلاةُ والسَلامُ في مَعْنى الآيَةِ، ثُمَّ أباحَ بَعْدُ لِمَعْنى الِاتِّعاظِ لا لِمَعْنى المُباهاةِ والِافْتِخارِ كَما يَفْعَلُ الناسُ في مُلازَمَتِها وتَسْنِيمِها بِالرُخامِ والحِجارَةِ، تَلْوِينُها سَرَفًا، وبُنْيانَ النَواوِيسِ عَلَيْها.
وقَوْلُهُ تَعالى: ﴿كَلا سَوْفَ تَعْلَمُونَ﴾ زَجْرٌ ووَعِيدٌ، ثُمَّ كَرَّرَ تَعالى: "كَلّا" تَأْكِيدًا، ويَأْخُذُ الناسَ مِن هَذا الزَجْرِ والوَعِيدِ المُكَرَّرَيْنِ كُلُّ أحَدٍ عَلى قَدْرِ حَظِّهِ مِنَ التَوَغُّلِ فِيما يَكْرَهُ، هَذا تَأْوِيلُ جُمْهُورِ الناسِ، وقالَ عَلِيُّ بْنُ أبِي

طالِبٍ رَضِيَ اللهُ عنهُ "كَلّا سَتَعْلَمُونَ في القُبُورِ، كَلّا سَتَعْلَمُونَ في البَعْثِ، وقالَ الضحّاكُ: الزَجْرُ الأوَّلُ وعِيدُهُ هو لِلْكُفّارِ والثانِي لِلْمُؤْمِنِينَ. وقَرَأ مالِكُ بْنُ دِينارٍ: "كَلّا سَيَعْلَمُونَ" فِيهِما.

In the name of Allah, the Most Gracious, the Most Merciful.

Interpretation of Surah At-Takathur:
And it is a Meccan surah, and I do not know of any disagreement regarding it.

Allah, exalted be He, says:
1 Competing for more distracts you 2 until you go into your graves. 3 No indeed! You will come to know. 4 No indeed! In the end you will come to know. 5 No indeed! If only you knew for certain! 6 You will most definitely see Hellfire. 7 Then you will see it with the eye of certainty. 8 On that Day, you will be asked about your pleasures.

"Alhakum" means: It has preoccupied you with its pleasures, and among them is "engaging in useless talk and voices" and frivolity with women. This is a statement that includes criticism, reproach, and regret. Ibn Abbas, Abu Imran al-Jawni, and Abu Salih recited "Alhakum" as an interrogative expression.
"And At-Takathur" means boasting about wealth, children, and abundance in general. This is the characteristic of the people of this world and its inhabitants, including the Arabs and others. Only the pious scholars are exempt from this. Al-A'sha said:

"I am not more than them in numbers,
But the honor belongs to the abundant."

The Prophet  said: "The son of Adam says: 'My wealth, my wealth!' Have you any wealth, O son of Adam, except what you have eaten and consumed, what you have worn and worn out, or what you have given in charity and sent forth?"

The interpreters differed regarding the meaning of Allah's statement: "Until you visit the graves." Some of them said: Until you mention the dead in your boasting about your ancestors and forefathers, and you have become numerous in your lifeless bones. Others said: The meaning is until you die and visit your graves with your bodies, meaning, you have cut short their lives due to your preoccupation with rivalry. It is narrated that an Arab heard this verse and said: "The people are sent to the Day of Judgment and the Lord of the Kaaba, for the visitor does not stay" and An-Naqqaash narrated this inclination from Umar ibn Abdul Aziz. Others said: This is a reproach for excessive visitation of graves, meaning, you have made your occupation, which diverts you from knowledge and learning, the excessive visitation of graves, glorifying and praising those who have passed away. And it is said: Then the Prophet  said, 'I used to forbid you from visiting the graves, but now visit them and do not abandon it.' His prohibition  was in the sense of the verse, then he allowed it later for the purpose of admonition, not for the purpose of boasting and showing off, as people do in their attachment and adoration of marble and stone, decorating it extravagantly and building domes over them.

And His statement, exalted be He: "Nay! You shall come to know!" is a reprimand and warning, and then He repeated it to emphasize it, saying "Nay!" as confirmation. People take from this repeated reprimand and warning according to their level of indulgence in what is disliked. This is the interpretation of the majority of people. Ali ibn Abi Talib, may Allah be pleased with him, said: "Nay! You shall come to know in the graves. Nay! You shall come to know in the resurrection." Ad-Dahhak said: The first reprimand and warning are for the disbelievers, and the second is for the believers. And Malik ibn Dinar recited in both instances, "Nay! They shall come to know."

### *Bahr Al 'Ulum by Abu al-Layth al-Samarqandi*

This book is considered one of the earliest works of interpretation based on transmitted narratives (tafsir bi'l-ma'thur). Its author skillfully combines both the approach of narrated interpretation and intellectual interpretation, although the emphasis leans more towards the transmitted aspect rather than the intellectual one. Consequently, it is classified among the books of interpretation relying on transmitted narratives. The author's methodology in this book primarily involves presenting narrations from the companions, their followers, and subsequent generations without extensively delving into the chains of narration. Whenever possible, the author elucidates the meaning of a verse by referring to other Quranic verses. Furthermore, the book incorporates certain Isra'iliyyat stories within its interpretation. This tafsir also addresses sequences of related verses that can sometimes include the entire surah. Example tafsir for surah 102:

مختلف فيها وهي ثمان آيات مكية

قوله تعالى أَلْهاكُمُ التَّكاثُرُ قال الكلبي نزلت في حَيَّيْن من العرب أحدهما بنو عبد مناف والآخر بنو سهم تفاخرا في الكثرة فكثرتهم بنو عبد مناف فقال بنو سهم إنا البغي والقتال قد أهلكنا فقد أحيانا وأحياكم وأمواتنا وأمواتكم ففعلوا فكثرتهم بنو سهم فنزل (أَلْهَاكُمُ التكاثر) يعني: شغلكم وأذهلكم التفاخر حَتَّى زُرْتُمُ الْمَقابِرَ يعني: أتيتم وذكرتم وعددتم أهل المقابر يعني: حتى يدرككم الموت على تلك الحال وروي عن النبي أنه قرأ أَلْهاكُمُ التَّكاثُرُ حَتَّى زُرْتُمُ الْمَقابِرَ ثم قال يقول بني آدم مالي مالي وهل لك من مالك إلا ما أكلت فَأَفْنَيْتَ أو لبست فأبليت أو تصدقت فأمضيت ويقال معناه أغفلكم التفاخر والتكاثر عن الهاوية والنار الحامية حتى زرتم المقابر يعني: عددتم مَنْ في المقابر ثم قال كَلَّا وهو رد على صنيعكم ويقال (كلا) معناه أي لاَ تَدَعون الفخر بالأحساب حتى زرتم المقابر وقال الزجاج كلا ردع لهم وتنبيه يعني: ليس الأمر الذي أن يكون عليه التكاثر والذي ينبغي أن يكونوا عليه طاعة الله تعالى والإيمان بنبيه محمد سَوْفَ تَعْلَمُونَ إذا نزل بكم الموت ويقال (كلا سوف تعلمون) إن سئلتم في القبر ثم قال ثُمَّ كَلَّا سَوْفَ تَعْلَمُونَ بعد الموت حين نزل بكم العذاب لأن الأحساب لا تنفعكم قوله تعالى كَلَّا لَوْ تَعْلَمُونَ قال بعضهم معناه لا لا تؤمنون بالوعيد وقد تم الكلام ثم استأنف فقال عِلْمَ الْيَقِينِ يعني: لو تعلمون ما القيامة باليقين لألهاكم عن ذلك ويقال هذا موصول به كلا لو تعلمون يقول حقاً لو علمتم علم اليقين بأن المال والحسب والفخر لا ينفعكم يوم القيامة ما افتخرتم بالمال والعدد والحسب ثم قال عز وجل لَتَرَوُنَّ الْجَحِيمَ.

قرأ ابن عامر والكسائي لَتَرَوُنَّ بضم التاء والباقون بالنصب فمن قرأ بالضم فهو على فعل ما لم يسم فاعله ونصب الجحيم على أنه مفعول به ثان، ومن قرأ بالنصب فعلى فعل المخاطبة ونصب الجحيم لأنه مفعول يعني: لترون الجحيم يوم القيامة عياناً ثُمَّ لَتَرَوُنَّها عَيْنَ الْيَقِينِ يعني: يدخلونها عياناً لا شك فيه ثُمَّ لَتُسْئَلُنَّ يَوْمَئِذٍ عَنِ النَّعِيمِ يعني: ولتسألن يوم القيامة عن النعيم قال علي بن أبي طالب من أكل خبزاً يابساً وشرب الماء من الفرات فقد أصاب النعيم وقال ابن مسعود هو الأمن والصحة وروى حماد بن سلمة عن أبيه عمار بن أبي عمار عن جابر أنه قال جاءنا رسول الله وأبو بكر وعمر ما فأطعمناهم رطباً وأسقيناهم الماء فقال رسول الله «هذا من النعيم الذي تَسْأَلُونَ عَنْهُ» وروى صالح بن محمد عن محمد بن مروان عن الكلبي عن أبي صالح

عن ابن عباس قال إن أبا بكر سأل رسول الله عن أكلة أكلها مع رسول الله في بيت أبي الهيثم بن التيهان من لحم وخبز وشعير وبسر مذنب وماء عذب فقال لرسول الله : أتخاف علينا أن يكون هذا من النعيم الذي نسأل عنه فقال النبي : «إِنَّمَا ذلك لِلكُفَّارِ ثُمَّ قَالَ ثَلاَثَةٌ لا يَسْأَلُ الله تَعالَى عَنْهَا العَبْدَ يَوْمَ القِيَامَةِ مَا يُوارِي عَوْرَتَهُ وَمَا يُقِيمُ بِهِ صُلْبُهُ وَمَا يَكُفُّهُ عَنِ الحَرِّ والقُرِّ وَهُوَ مَسْؤُولٌ بَعْدَ ذلك عَنْ كُلِّ نِعْمَةٍ» وروى الحسن عن رسول الله أنه قال «مَا أَنْعَمَ الله تَعَالَى عَلَى العَبْدِ مِنْ نِعْمَةٍ صَغِيرَةٍ أوْ كَبِيرَةٍ فَيَقُولُ عَلَيْهَا الحَمْدُ لله إِلاَّ أَعْطَاهُ الله تَعَالى خَيْراً مِمَّا أَخَذَ» والله أعلم وعن رسول الله أنه قال: «من قرأ سُورَةِ التَّكَاثُر لَمْ يُحَاسِبْهُ الله تَعَالَى بِالنَّعِيمِ الَّذِي أَنْعَمَ بِهِ فِي الدَّارِ الدُّنْيَا وَأُعْطِيَ مِنَ الأَجْرِ كأنّما قرأ القرآن» .

Different opinions exist regarding it, and it consists of eight Meccan verses.
Allah, the Most High, says: "The mutual rivalry diverts you." Al-Kalbi said: It was revealed regarding two tribes among the Arabs, one of them being the Banu Abd Manaf, and the other being the Banu Sahm, who boasted about their abundance. The descendants of 'Abd Manaf boasted about their numbers, so the descendants of Sahm said, "In the era of ignorance, injustice destroyed us, and they retaliated against us with the living and the dead," so the descendants of Sahm boasted about their numbers, so the verse (The mutual rivalry) was revealed, meaning: It preoccupied you and amazed you, boasting about your abundance, until you visit the graves, meaning: You come and mention and count the people in the graves, meaning: Until death overtakes you in that state. It is narrated from the Prophet that he recited the verse (The mutual rivalry) until you visit the graves. Then he said: The son of Adam says, "My wealth, my wealth!" Have you any wealth, O son of Adam, except what you have eaten and consumed, what you have worn and worn out, or what you have given in charity and sent forth? It is also said that its meaning is: It distracted you from boasting and rivalry about the abyss and the blazing fire until you visit the graves, meaning: You counted those who are in the graves. Then Allah, exalted be He, said: Nay! And this is a response to your actions. And it is said that (Nay!) means: No, do not boast about the accounts until you visit the graves. And Az-Zajjaj said: Nay! is a deterrent and a warning, meaning: The matter is not as you think it is. It should be that you obey Allah, exalted be He, and have faith in His Prophet Muhammad. You will surely come to know when death descends upon you. It is also said: Nay! You shall come to know if you are asked in the grave. Then He said: Then, on that Day, you shall be asked about the delights. Meaning: You will be asked on the Day of Judgment about the blessings. Ali ibn Abi Talib said: Whoever eats dry bread and drinks water from the Euphrates has attained the blessings (the delights). Ibn Mas'ud said: It is security and good health. Hammad ibn Salamah narrated from his father, from Ammar ibn Abi Ammar, from Jabir, who said: The Messenger of Allah, Abu Bakr, and Umar came to us. We fed them dates and gave them water. The Messenger of Allah said, "This is from the blessings that you will be asked about." Salih ibn Muhammad narrated from Muhammad ibn Marwan, from Al-Kalbi, from Abu Salih, from Ibn Abbas, who said: Abu Bakr asked the Messenger of Allah about a meal that he had eaten with the Messenger of Allah in the house of Abu al-Haytham ibn al-Tihhan, which consisted of meat, bread, barley, and vegetables, along with pure water. Abu Bakr asked the Messenger of Allah, "Do you fear for us that this is from the blessings that we will be asked about?" The Prophet said, "Indeed, that is for the disbelievers." Then he said, "Three things that a servant will not be asked about on the Day of Judgment: what he covers his private parts with, what he fulfills his needs with, and what he shields himself from the heat and cold with. But he will be asked about all the blessings after that." Hasan narrated from the Messenger of Allah that he said, "Whenever a servant of Allah recites Surah Al-Takathur, he will not be called to account for the blessings that he was given in this worldly life, and he will be given the reward as if he had recited the entire Qur'an."

# Experiments

In this study, we propose a framework that integrates multiple models and Qur’an tafsirs, as depicted in Figure 1. The first phase of the framework involves selecting one model at a time, which is then utilized to perform Word2Vec embedding on all Qur’an tafsirs. The Word2Vec architecture used here is Continuous Bag of Words (CBOW), which learns the probability distribution of a target word given its context. This process generates a tafsir tensor for each tafsir, and the tensor is indexed using the same indexing as the corresponding CSV file, where each ayah typically corresponds to one tafsir. Similarly, the prompt is converted into a vector using the same process. To find the most similar tafsir in the tafsir tensor, cosine similarity is employed. The index of the closest tafsir is used as a reference to retrieve the corresponding ayah from the respective Qur’an tafsir file.

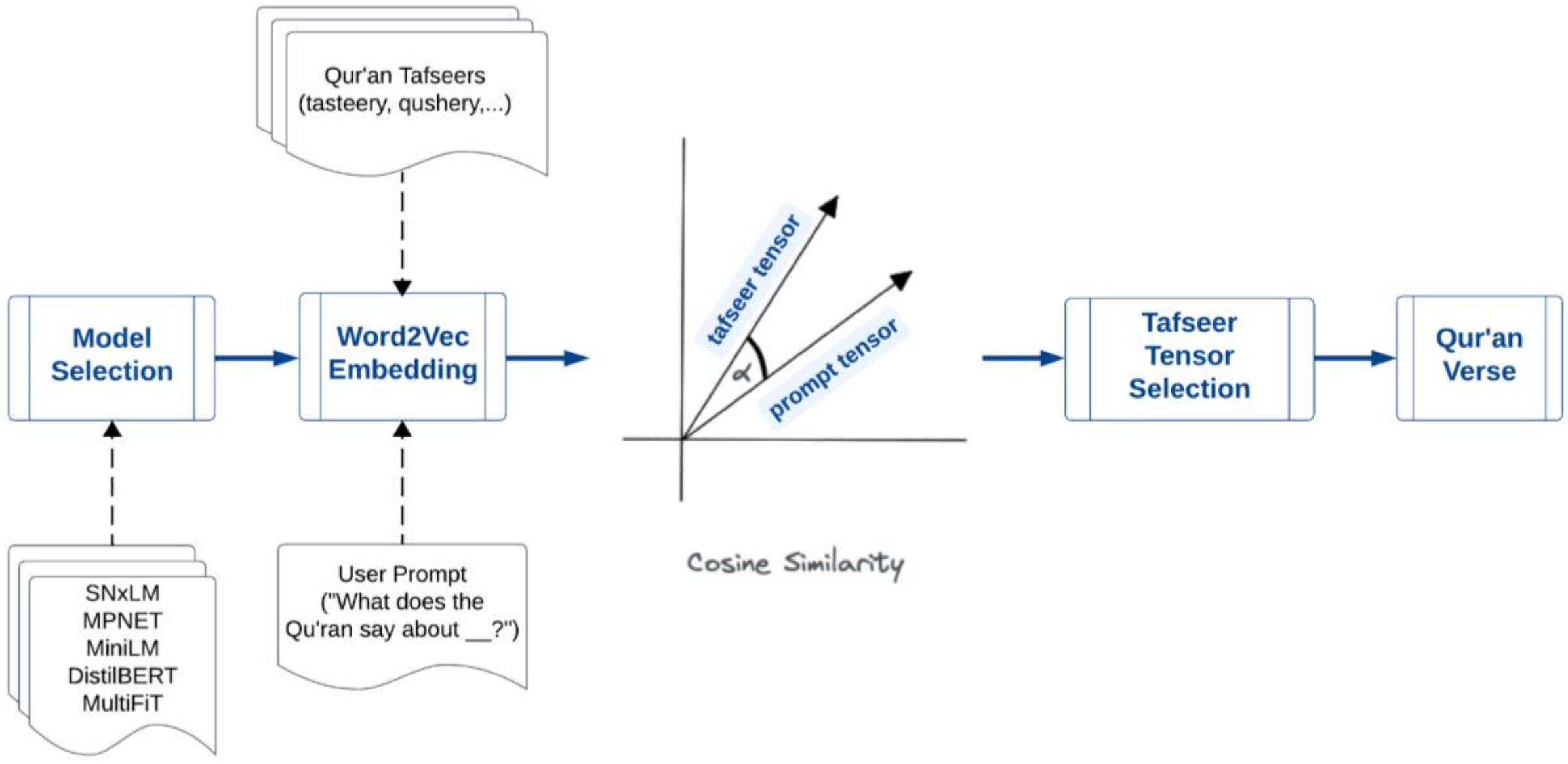

Figure 1: Overview of the proposed methodology

To process and organize the input tafsir files that exist in a JSON format, several modifications needed to be made. We first iterate through folders and JSON files containing Qur’anic verses, extract the verses' unique identifiers and corresponding explanations, fetch the text of those verses from a Quran dataframe, and save the extracted data as CSV files for each folder.
For each file, we initialize two empty lists: “tafsir_sid_div_vid” and “tafsir_list”. These lists will store information related to the verses and their corresponding explanations. We then iterate through a range of numbers from 1 to 115, representing the Surah number. Inside the loop, we format the verse number with leading zeros to match the filename pattern to check if a specific JSON file exists. If the file exists, it is loaded as a JSON object and assigned to the variable verse. The code then iterates through the data within the verse object. For each entry, it checks the verse reference “*vref*” to determine if it represents a single verse or a range of verses. If it's a single verse (no '-' exists), the verse's unique identifier (“*sid_div_vid*”) is calculated by raising the chapter number and adding the verse number. This unique identifier is added to the “*tafsir_sid_div_vid*” list, and the corresponding tafsir (explanation/commentary) is added to the “*tafsir_list*”. If it's a range of verses (indicated by '-' in ayah_check), the code determines the starting verse and ending verse (smaller and larger, respectively) and calculates the number of repeats. It then enters a loop that iterates through each verse in the range, calculates the unique identifier, and appends it to “*tafsir_sid_div_vid*” along with the corresponding tafsir. After processing all the verses within the current folder, the code creates an empty list called “*ayah_list*”. The code iterates through the “*tafsir_sid_div_vid*” list, and for each unique identifier, it filters the Qur’an dataframe (Quran_df) to retrieve the corresponding verse's text (ptext). The verse's text is appended to the “*ayah_list*”. Using the “*ayah_list*” and “*tafsir_list*”, a Pandas DataFrame (df) is created with two columns: "Ayah" (containing the verse text) and "*tafsir*" (containing the tafsir). Finally, the DataFrame df is saved as a CSV file with the name of the current folder in a directory called "*tafsir_csv*".

## Model Selection

Researchers and practitioners often choose a model based on their specific requirements, such as task complexity, computational resources and the availability of pre-trained models for a particular language or domain. The following models have been chosen because they address specific challenges in natural language processing (NLP) and have shown promising results across a range of tasks:

- SNxLM (Self-Normalizing Cross-Lingual Language Model): SNxLM is a language model that focuses on cross-lingual understanding and representation learning. It utilizes self-normalization techniques to effectively handle language variations and complexities across different languages. By leveraging large-scale multilingual data, SNxLM aims to learn language-agnostic representations that capture universal linguistic patterns and can be applied to various natural language processing tasks.
- MPNet (Masked and Permuted BERT): MPNet is an extension of the BERT (Bidirectional Encoder Representations from Transformers) model that incorporates additional pre-training objectives. In addition to the original masked language modeling, MPNet introduces permuted language modeling, where word order is randomly permuted within a sentence. This helps the model to capture more fine-grained dependencies and structural information. MPNet enhances the contextual understanding and representation learning capabilities of BERT, leading to improved performance on a wide range of natural language understanding tasks.

- MiniLM: MiniLM is a compact variant of the BERT model designed to be computationally efficient while maintaining strong language understanding capabilities. It achieves a smaller model size by reducing the number of layers, hidden units, and attention heads compared to the original BERT. Despite its smaller size, MiniLM utilizes advanced training techniques and architectural modifications to preserve the core principles of the Transformer model and achieve competitive performance on various NLP tasks.
- DistilBERT: DistilBERT is a distilled version of the BERT model that aims to compress the original model while retaining most of its performance. It is trained using a teacher-student knowledge distillation approach, where a larger BERT model serves as the teacher and transfers its knowledge to the smaller DistilBERT student model. By distilling the knowledge from the teacher model, DistilBERT achieves a reduced model size and faster inference time while preserving a significant portion of the original BERT's language representation capabilities.
- MultiFiT: MultiFiT is a language model that focuses on multilingual representation learning. It is trained on data from multiple languages simultaneously, allowing it to capture shared linguistic properties across different languages. By leveraging the similarities and transferable knowledge among languages, MultiFiT aims to create a unified representation space that can handle multiple languages effectively. This enables the model to provide contextualized embeddings and facilitate cross-lingual understanding, enabling applications in multilingual natural language processing tasks. These models are designed with specific objectives and techniques to address various challenges in language understanding, representation learning, and cross-lingual applications. They contribute to the advancement of natural language processing by providing more efficient, compact, and versatile models for processing textual data in different languages.

## Evaluation Metrics

Cosine similarity is a powerful measure employed in various fields, including NLP, to determine the similarity between textual data. In the context of analyzing the Qur'an, cosine similarity can be effectively utilized to identify the closest verse based on its content. To accomplish this, the verses are first transformed into numerical representations, such as word embeddings or TF-IDF vectors, which capture their semantic meaning. By calculating the cosine similarity between these vector representations, the degree of similarity between verses can be measured. A higher cosine similarity score indicates a stronger resemblance in the meaning and context of the verses. This approach enables researchers and scholars to efficiently explore and analyze the relationships between different Qur'an passages, identify related verses, and uncover patterns and themes within the text. By leveraging cosine similarity, the process of finding the closest verse in the Qur'an becomes a computationally efficient and data-driven task, facilitating a deeper understanding of the textual content and its interconnections. The formula for computing cosine similarity between two vectors A and B is as follows:

$$cosine\ similarity\ (A, B) = \frac{A \cdot B}{\| A \| \| B \|}$$

After converting Arabic tokens to numerical representations using Word2Vec, the numerical vectors undergo a normalization process to eliminate any bias that may arise from the document. Subsequently, the dot product of the normalized vectors is calculated. The dot product is obtained by summing the element-wise multiplication between the corresponding components of the vectors. To obtain the cosine similarity, the dot product is divided by the product of the magnitudes (or Euclidean norms) of the vectors. The resulting value, ranging from -1 to 1, serves as a measure of the cosine similarity between the two vectors. A value close to 1 indicates a high degree of similarity, whereas a value close to -1 signifies dissimilarity between the vectors.

# Results and Discussion

In this section we analyze the performance of each tafsir and model based on the accuracy and acceptability of results from different tafsirs. The accuracy refers to the number of instances where the model's output aligns with the expected result for a specific tafsir. The acceptability indicates the number of instances where the model's output is considered satisfactory. The results vary across the models and tafsir. For example, Alkashshaf received accurate results in one instance from SNxLM, while Altasheel had three accurate results from SNxLM and two acceptable results from distilMulti. SNxLM excels among the models, showcasing the highest count of accurate results, notably 3 for 'Altasheel'. However, the model lacks accurate results for some tafsirs like 'Matureedi', 'Zad-almaseer', and 'Roohalbayan'. Some tafsirs had no accurate or acceptable results from certain models.

***Experimental Results***

The following tables present a summary of the SNxLM model output for a subset of the prompts. For the topic of financial relations, Ibn Atiyah's interpretation (ibn-atiyah.csv) has a cosine similarity score of 0.44. The corresponding verse advises not to approach the wealth of orphans except in a manner that is best for them, and to fulfill measures and weights with justice, while also emphasizing not burdening anyone beyond their capacity and promoting fairness and justice. Samarqandi's interpretation (samarqandi.csv) has a cosine similarity score of 0.53. The corresponding verse emphasizes not approaching the wealth of orphans except in the best way possible and fulfilling agreements, as the fulfillment of agreements is a responsibility. Abdu's interpretation (abdu.csv) has a high cosine similarity score of 0.97. The corresponding verse addresses the believers and advises them not to consume their wealth among themselves unjustly unless it is through legitimate trade based on mutual consent. It also prohibits killing oneself, as God is merciful towards them. The verses that correspond to the interpretations mentioned earlier all explicitly use the term "money" and offer direct guidance on financial matters which accounts for their high actual relevancy score. Alkashshaf's interpretation (alkashaf.csv) has a cosine similarity score of 0.70. The corresponding verse advises believers not to give their wealth to those who are foolish and extravagant, but to provide for them in a reasonable manner and speak to them with kind words. These interpretations highlight the importance of fair treatment, fulfilling agreements, avoiding unjust consumption of wealth, and providing for those in need while exercising wisdom and moderation. Despite the presence of the term "money" in this verse, it received a medium actual relevancy score due to the relatively indirect guidance it provides on financial matters.

| Topic: Financial Relations | | | |
|---|---|---|---|
| **Tafsir** | **Cosine Similarity** | **Result** | **Actual Relevancy** |
| ibn-atiyah<br><br>*Muharar al-Wajiz* | 0.44 | ['ولا تقربوا مال اليتيم إلا بالتي هي أحسن حتى يبلغ أشده وأوفوا الكيل والميزان بالقسط لا نكلف نفسا إلا وسعها وإذا قلتم فاعدلوا ولو كان ذا قربى وبعهد الله أوفوا ذلكم وصاكم به لعلكم تذكرون'] | High |
| samarqandi<br><br>Bahr Al 'Ulum | 0.53 | ['ولا تقربوا مال اليتيم إلا بالتي هي أحسن حتى يبلغ أشده وأوفوا بالعهد إن العهد كان مسئولا'] | high |
| abdu | 0.97 | ['يا أيها الذين آمنوا لا تأكلوا أموالكم بينكم بالباطل إلا أن تكون تجارة عن تراض منكم ولا تقتلوا أنفسكم إن الله كان بكم رحيما'] | high |
| alkashaf<br><br>Al-Kashshaaf 'an Haqa'iq at-Tanzil | 0.70 | ['ولا تؤتوا السفهاء أموالكم التي جعل الله لكم قياما وارزقوهم فيها واكسوهم وقولوا لهم قولا معروفا'] | medium |

Table 1: Tafsir, cosine similarity, corresponding verse and actual relevancy of the verse to the topic of “Financial Relations”.

For the topic of forgiveness, Altasheel's interpretation (altasheel.csv) has a cosine similarity score of 0.59. The corresponding verse advises to embrace forgiveness, promote good conduct, and turn away from the ignorant. Due to the explicit use of the word “forgiveness” and the direct advice on the matter, this verse received a high actual relevancy score.

| **Topic: Forgiveness** | | | |
|---|---|---|---|
| altasheel | 0.59 | ['خذ العفو وأمر بالعرف وأعرض عن الجاهلين'] | High |

Table 2: Tafsir, cosine similarity, corresponding verse and actual relevancy of the verse to the topic of “Forgiveness”.

For the topic of violence, Altasheel's interpretation (altasheel.csv) has a cosine similarity score of 0.93. The corresponding verse permits fighting in the cause of God against those who fight against Muslims, but it prohibits transgression as God does not love those who transgress. The corresponding verse received a high actual relevancy score as it not only featured the term "fight" multiple times but also contained a direct command from God. Aldorr's interpretation (aldorr.csv) has a cosine similarity score of 0.60. The corresponding verse mentions that fighting has been prescribed for Muslims even though they may dislike it. However, what they dislike may be good for them, and what they like may be harmful as God has knowledge that humans do not possess. Tabari's interpretation (tabari.csv) has a cosine similarity score of 0.63. The corresponding verse grants permission to fight against those who have wronged Muslims and emphasizes that God is capable of supporting them in their fight for justice. These interpretations highlight the significance of forgiveness, good conduct and the permissibility of fighting in self-defense or in defense of one's faith. They also stress the importance of avoiding transgression and seeking God's support in matters of violence. The above two verses received an actual relevancy score of medium because, relative to the first verse, the word “fight” was only mentioned once and the verses are written in a passive rather than an active order.

| **Topic: Violence** | | | |
|---|---|---|---|
| altasheel | 0.93 | ['وقاتلوا في سبيل الله الذين يقاتلونكم ولا تعتدوا إن الله لا يحب المعتدين'] | High |
| aldorr | 0.60 | ['كتب عليكم القتال وهو كره لكم وعسى أن تكرهوا شيئا وهو خير لكم وعسى أن تحبوا شيئا وهو شر لكم والله يعلم وأنتم لا تعلمون'] | Medium |
| tabari | 0.63 | ['أذن للذين يقاتلون بأنهم ظلموا وإن الله على نصرهم لقدير'] | Medium |

Table 3: Tafsir, cosine similarity, corresponding verse and actual relevancy of the verse to the topic of "Violence".

For the topic of women, Tabari's interpretation (Tabari.csv), Almokhtasar's interpretation, and Alaloosi's interpretation share the same cosine similarity score of 0.53. The corresponding verses instruct believers to give women their bridal gifts willingly, and if they willingly offer a portion of it to the husband, then they may consume it happily and with pleasure. The corresponding verses received a high score for actual relevancy due to its direct advice on how to treat women in Islam. Zad-Almasser's interpretation has a cosine similarity score of 0.52. The verse begins with "In the name of Allah, the Most Gracious, the Most Merciful" and addresses all people, urging them to be mindful of their Lord who created them from a single soul and created from it its mate. It emphasizes the importance of fearing God and maintaining family ties, as God is watchful over them. Alkashaf's interpretation has a cosine similarity score of 0.59. The verse states that there is no blame on believers if they divorce women without having touched them or fixed a dowry for them. However, they should provide for them according to their means and treat them kindly, offering them a suitable provision according to what is reasonable and customary. Qusheeri's interpretation has a cosine similarity score of 0.42. The verse refers to Maryam, the daughter of Imran, who guarded her chastity, and God blew His spirit into her. It acknowledges her truthfulness in accepting the words of her Lord and His scriptures and highlights her devoutness and obedience. While the three verses mentioned above do include the term "women" or reference a woman (Maryam), their length dilutes the presence of these words within the verses, potentially accounting for the actual relevancy score of medium.

| **Topic: Women** | | | |
|---|---|---|---|
| Tabari | 0.53 | ['وآتوا النساء صدقاتهن نحلة فإن طبن لكم عن شيء منه نفسا فكلوه هنيئا مريئا'] | High |
| Almokhtasar | 0.53 | ['وآتوا النساء صدقاتهن نحلة فإن طبن لكم عن شيء منه نفسا فكلوه هنيئا مريئا'] | High |
| Alaloosi | 0.53 | ['وآتوا النساء صدقاتهن نحلة فإن طبن لكم عن شيء منه نفسا فكلوه هنيئا مريئا'] | High |
| Zad-almasser | 0.52 | ['بسم الله الرحمن الرحيم يا أيها الناس اتقوا ربكم الذي خلقكم من نفس واحدة وخلق منها زوجها وبث منهما رجالا كثيرا ونساء واتقوا الله الذي تساءلون به والأرحام إن الله كان عليكم رقيبا'] | Medium |
| Alkashaf | 0.59 | ['لا جناح عليكم إن طلقتم النساء ما لم تمسوهن أو تفرضوا لهن فريضة ومتعوهن على الموسع قدره وعلى المقتر قدره متاعا بالمعروف حقا على المحسنين'] | Medium |
| Qusheeri | 0.42 | ['ومريم ابنت عمران التي أحصنت فرجها فنفخنا فيه من روحنا وصدقت بكلمات ربها وكتبه وكانت من القانتين'] | Medium |

Table 4: Tafsir, cosine similarity, corresponding verse and actual relevancy of the verse to the topic of "Women".

For the topic of good behaviors, Baghawi's interpretation (Baghawi.csv) has a cosine similarity score of 0.30. The corresponding verse states that those who believe in Allah and the Day of Judgment, enjoin what is right, forbid what is wrong, and hasten to do good deeds are among the righteous. The actual relevancy of this verse is high due to the presence of the word "goodness" in the verse. Almokhtaser's interpretation (Almokhtaser.csv) has a cosine similarity score of 0.37. The corresponding verse mentions that those who listen to the word of God and follow the best of it are the ones whom Allah has guided, and they possess understanding. Ibn Kathir's interpretation (Ibn-katheer.csv) has a cosine similarity score of 0.49. The verse mentioned in this interpretation advises not to be deceived by the arguments of those who disbelieve in the signs of Allah. Their changing situations in different places should not distract or mislead believers. These interpretations highlight good behaviors such as belief in Allah and the Day of Judgment, enjoining righteousness, forbidding wrongdoing, hastening to do good deeds, listening to the word of God, following the best of it, and not being deceived by the arguments of disbelievers. Although the above two verses entail what is regarded as good behavior, they do not possess the word “good” or any form of it, which may have contributed to the medium score of actual relevance.

| **Topic: Good Behaviors** | | | |
|---|---|---|---|
| Baghawi | 0.30 | ['يؤمنون بالله واليوم الآخر ويأمرون بالمعروف وينهون عن المنكر ويسارعون في الخيرات وأولئك من الصالحين'] | High |
| Almokhtaser | 0.37 | ['الذين يستمعون القول فيتبعون أحسنه أولئك الذين هداهم الله وأولئك هم أولو الألباب'] | Medium |
| Ibn-katheer | 0.49 | ['ما يجادل في آيات الله إلا الذين كفروا فلا يغررك تقلبهم في البلاد'] | Medium |

Table 5: Tafsir, cosine similarity, corresponding verse and actual relevancy of the verse to the topic of “Good Behaviors”.

For the topic of the role of knowledge, Alkashaf's interpretation (Alkashaf.csv) has a cosine similarity score of 0.68. The corresponding verse calls upon people to reflect upon the creation of the heavens, the earth, and what is between them. It emphasizes that all of this has been created with truth and for a specified term. However, many people are ungrateful disbelievers when it comes to meeting their Lord. Altasheel's interpretation (Altasheel.csv) has a cosine similarity score of 0.39. The verse mentioned in this interpretation highlights that those who have been granted knowledge from their Lord recognize it as the truth and it guides them to the path of the Mighty, the Praiseworthy. These interpretations emphasize the importance of knowledge in Islam. They indicate that knowledge leads to reflection, recognition of truth, and guidance towards the path of God. Knowledge is seen as a means to gain deeper understanding, appreciate the signs of creation, and align oneself with the divine path. Both verses received an actual relevancy score of medium, which could be due to the rare occurence of the word “knowledge” in the verses.

| **Topic: The Role of Knowledge** | | | |
|---|---|---|---|
| Alkashaf | 0.68 | ['أولم يتفكروا في أنفسهم ما خلق الله السماوات والأرض وما بينهما إلا بالحق وأجل مسمى وإن كثيرا من الناس بلقاء ربهم لكافرون'] | Medium |
| Altasheel | 0.39 | ['ويرى الذين أوتوا العلم الذي أنزل إليك من ربك هو الحق ويهدي إلى صراط العزيز الحميد'] | Medium |

Table 6: Tafsir, cosine similarity, corresponding verse and actual relevancy of the verse to the topic of “The Role of Knowledge".

## Conclusion & Future Work

This paper introduces a framework for constructing a semantic search tool tailored to the Qur’an, allowing users to effectively identify all relevant verses pertaining to a particular topic, usually filling the blank to the question “What does the Qur’an say about __ ?”. To achieve this, the process involves selecting one model at a time from the list of chosen models (SNxLM, MPNet, MiniLM,

DistilBERT or MultiFiT) and using it to create Word2Vec embeddings for all Qur'anic tafsirs. These embeddings result in a tafsir tensor for each tafsir, where each tafsir corresponds to one ayah (verse) and vector for the input prompt. Finally, cosine similarity is the distance metric of choice, which calculates the cosine of the angle between these vectors to indicate how closely they are aligned. The index of the tafsir that is closest to the prompt vector is then used to retrieve the corresponding ayah that is relevant to the prompt. Results show that SNxLM excels among the models as it showcases the highest count of accurate results for the particular tafsir called 'Altasheel'. One of the highest cosine similarity scores of 0.97 corresponds to the "abdu" tafsir for a verse relating to financial matters. In the future, we plan to experiment with different similarity measures, other than cosine similarity, that could potentially obtain deeper semantic-based similarity. We also plan to experiment with other popular word embedding models used in NLP and text analysis such as GloVe or FastText and other large language models such as LLaMa from Meta [15] and LaMDA from Google [16] that may potentially improve search accuracy by understanding the intent and contextual meaning of terms.

## References

[1] "The Future of the Global Muslim Population," *Pew Research Center's Religion & Public Life Project* (blog), January 27, 2011, https://www.pewresearch.org/religion/2011/01/27/the-future-of-the-global-muslim-population/.

[2] Ingrid Mattson, *The Story of the Qur'an: Its History and Place in Muslimlife*, 2nd ed (Chichester, West Sussex ; Malden, MA: Wiley-Blackwell, 2013), 146.

[3] Jerald Dirks, *The Cross & The Crescent*, 1st ed (Beltsville, Md: Amana Publications, 2001), 45.

[4] M. A. S. Abdel Haleem, trans., *The Qur'an: English Translation and Parallel Arabic Text*, Bilingual, Revised edition (Oxford ; New York: Oxford University Press, 2010), 7.

[5] Haleem, 604.

[6] "Lanes Lexicon," accessed May 20, 2023, http://lexicon.quranic-research.net/data/06_H/123_HTb.html.

[7] "Hafs Narration," King Fahd Glorious Qur'an Complex, September 29, 2019, https://qurancomplex.gov.sa/en/riwaiat-hafs/.

[8] SearchTruth, "Complete Indexing of Quran Topics from A to Z | SearchTruth," SearchTruth.com, accessed May 21, 2023, https://www.searchtruth.com/quran/topics/.

[9] "The Quranic Arabic Corpus - Topic Index," accessed May 21, 2023, https://corpus.quran.com/topics.jsp.

[10] Qortoba Foundation, "Quran Tafsirs," HTML, March 15, 2022, https://github.com/qortoba/tafsirs.

[11] "Tafsir Al Jalalayn Details Page," accessed May 21, 2023, https://www.quran.link/books/details/70/.

[12] "Al Tafsir Al Kabir Details Page," accessed May 21, 2023, https://www.quran.link/books/details/69/.

[13] M. A. S. Abdel Haleem, trans., The Qur'an: English Translation and Parallel Arabic Text, Bilingual, Revised edition (Oxford ; New York: Oxford University Press, 2010), 601.

[14] https://chat.openai.com/, accessed August, 2023.

[15] https://ai.meta.com/llama/

[16] https://blog.google/technology/ai/lamda/